\documentclass[11pt]{article}

\usepackage[preprint]{acl}

\usepackage{times}
\usepackage{latexsym}
\usepackage{booktabs}
\usepackage[T1]{fontenc}
\usepackage[utf8]{inputenc}
\usepackage{microtype}
\usepackage{inconsolata}
\usepackage{amsmath}
\usepackage{graphicx}
\usepackage{tikz}
\usetikzlibrary{
  positioning,
  shapes,
  arrows.meta,
  fit,
  backgrounds
}
\usepackage{xcolor}
\usepackage{tabularx}
\usepackage{makecell}
\usepackage{cleveref}
\usepackage{array}
\usepackage{caption}
\usepackage{subcaption}
\usepackage[section]{placeins}

\newcolumntype{L}{>{\raggedright\arraybackslash}X}
\newcolumntype{C}{>{\centering\arraybackslash}X}
\newcolumntype{R}{>{\raggedleft\arraybackslash}X}

\title{Natural-Language Policies to Executable Decisions: An Interpretable Large Language Model Framework}

\author{
{\textbf{Ziqiang Zhang}},
{\textbf{Jing Ma}},
{\textbf{Zilong Wang}},
{\textbf{Jiayuan Chen}},
{\textbf{Yi Qiao}},
{\textbf{Yu He}},  \\
{\textbf{Wei Zhang}},
{\textbf{Dai Cheng}},
{\textbf{Xiaoyu Shen}}\thanks{Corresponding author.} \\
Ningbo Institute of Digital Twin, Eastern Institute of Technology, Ningbo \\
{\tt zqzhang@idt.eitech.edu.cn \quad xyshen@eitech.edu.cn}
}

\allowdisplaybreaks  
\begin{document}
\maketitle
\begin{abstract}
Pricing automation in large-scale tourism is challenging because travel orders are highly unstructured, while pricing policies are complex, rapidly evolving, and inherently open-ended. Traditional rule engines are brittle and costly to maintain, whereas unconstrained LLM agents lack the reliability and auditability required for financial decisions. We present a production-grade LLM-powered pricing system with a strict decision boundary: LLMs perform structured extraction and bounded policy/path selection, while all numeric pricing, including total-price computation, is executed deterministically. Policies are compiled into interpretable condition trees, enabling open-ended support for new clauses and evolving rules without code changes, while exposing auditable artifacts for human-in-the-loop control. Periodic fine-tuning on logged traces further improves tree induction and path matching. Deployed at a municipal state-owned tourism enterprise across 7 scenic sites and 12 business categories with 1{,}500+ operators and 1{,}000+ active policies, the system processed 3{,}960 orders in six months, reduced the order management team from 15--20 to 3, and cut per-order handling time from $\sim$10 minutes to $<2$ minutes.
\end{abstract}

\section{Introduction}
Accurate and timely processing of travel orders is the operational core of modern tourism services. A typical order describes a group's itinerary, including scenic spots, travel dates, agent name, etc, which must be transformed into structured facts and matched with complex repository of pricing policies~\cite{Kaushik2017MakingTS}. While automating this workflow is essential for operational scalability, it remains remarkably difficult in practice due to the unstructured nature of travel requests and the intricate, ever-evolving pricing policies~\cite{zhou2025rulearena,Liu2025PricingLogicEL}.

In real-world production settings, travel orders exhibit extreme heterogeneity. Requests range from semi-structured digital forms to casual instant-messaging text or even handwritten notices. Critical information is often implicit or ambiguous: scenic spots may be referred to by informal abbreviations, and travel dates frequently rely on background context rather than explicit mention. These factors transform simple data parsing into a high-order information-understanding task that requires reasoning well beyond surface-level patterns \citep{Xu2019LayoutLMPO,Mathew2020DocVQAAD}.

This input complexity is compounded by the fundamentally \emph{open-ended} nature of pricing policies \citep{Hendrycks2021CUADAE}. Determining the applicability of a policy requires reasoning over a vast array of interacting conditions, such as seasonal windows, group size thresholds, and bespoke contractual clauses, distributed across heterogeneous policy documents. Because new policies routinely introduce novel conditions that were never anticipated at system design time, traditional software-based solutions that rely on manually encoding logic into structured databases and rule engines \citep{Desmond2022ANL} are inherently brittle. Each policy update often necessitates synchronized changes to backend schemas, rule code, and user interfaces, creating prohibitive maintenance costs and making city-scale deployment unscalable.

Recent advances in large language models (LLMs) offer a flexible alternative for reasoning over natural language~\cite{su2022welm,achiam2023gpt,liu2024deepseek,xu2025towards,ding2026llms}. In principle, LLMs can parse informal orders, interpret policy text, and perform cross-document reasoning without rigid schemas. In practice, however, fully autonomous LLMs are ill-suited for high-stakes applications. They often lack stability, provide limited interpretability, and offer no clear mechanism for non-technical users to inspect or correct intermediate reasoning steps, which is an essential requirement for pricing and auditability \citep{Agarwal2024FaithfulnessVP}.

In this work, we present a production-grade, LLM-powered pricing system that reconciles the linguistic flexibility of large language models with the strict reliability requirements of real-world financial operations. Our central insight is that LLMs should not replace pricing systems, but instead operate within a carefully defined and interpretable decision boundary. The system is designed around two core principles:
(1) \textit{Open-ended condition support}: Instead of hard-coding logic, we leverage LLMs to match order-side facts against policy-side condition trees. This design naturally supports an evolving rule space, enabling the system to handle arbitrary ad-hoc rules and seasonal exceptions without code changes or model retraining \citep{Gao2022PALPL,Chen2022ProgramOT}. (2) \textit{End-to-end interpretability}: The system automatically extracts logic from policy documents and organizes it into explicit \textit{condition trees}. These trees preserve the semantics of natural language while exposing their logical structure, allowing tourism managers to review and maintain rules without programming knowledge~\cite{xiong2024gptree,wang2025fault2flow}.
Crucially, all numeric computation is handled by a deterministic engine. LLMs are restricted to structured extraction and discrete decision selection. This separation provides the reliability of traditional software systems while retaining the reasoning flexibility of modern language models.

We deployed the system at a municipal, state-owned tourism enterprise operating a city-wide platform spanning seven major scenic areas and twelve business categories, with 1{,}500+ operators. Over the first six months after rollout, the system processed 3{,}960 orders, reduced the order-management team from 15--20 to 3, and cut per-order handling time from $\sim$10 minutes to $<2$ minutes. The deployment externalizes pricing governance into auditable artifacts---\texttt{OrderFacts}, policy-induced condition trees, and logged decision traces---making disagreements actionable via bounded human confirmation and overrides. These traces enable systematic error diagnosis and iterative improvements, without changing numeric execution. An internal survey reports improved communication/efficiency (96.92\%) and positive feedback on usability and analyzable rules (76.92\%).

While our deployment focuses on tourism, the challenges of open-ended policy logic, informal inputs, strict correctness requirements, and human-centered governance are common across many domains such as insurance underwriting, compliance checking, and contract execution. By eliminating rigid rule engineering while preserving human agency, our system offers a practical path toward scalable, trustworthy AI in policy-driven industries.

\section{Background}

\paragraph{The Pricing Lifecycle in Tourism}
Operational pricing in modern tourism follows a recurring tripartite lifecycle: (i) \textit{policy onboarding}, (ii) \textit{order intake and quotation}, and (iii) \textit{settlement and verification}. The process begins when business managers publish pricing policies—documents (often PDFs or spreadsheets) that define the rules for various customer segments and travel seasons. Once published, tour coordinators receive travel orders and must match them against the active policy repository to generate a quote. Finally, the calculated price must be verified against the original policy to ensure auditability and financial compliance.

\paragraph{Open-Ended Complexity of Pricing Policies}
A core challenge in this domain is that pricing conditions are fundamentally \textit{open-ended}. Policies are defined over \emph{resources} (atomic sellable units like a cable car ride) and \emph{products} (bundles of resources). However, the conditions governing their price, ranging from specific age brackets and group size thresholds to complex seasonal overlaps and ad-hoc contractual clauses, are virtually unlimited. New policies frequently introduce entirely novel logic that was not anticipated during initial system design. This evolving complexity means that ``pricing logic'' is not a static set of parameters, but a growing library of natural-language rules that must be interpreted in context.

\paragraph{Informality and Mismatch in Travel Orders}
The intake side of the lifecycle is equally challenging due to the extreme informality of travel orders. In our deployment, 97\% of orders (based on 2025H2 production logs) are received as instant-messaging screenshots rather than structured digital forms. These  requests mix travel dates, destinations, and casual notes without a fixed template. Furthermore, a ``resource mismatch'' is common: the itinerary described in the initial text often deviates from the final executed plan. Nearly half of our production cases require manual edits to the item list before a quotation can be finalized, necessitating a system that can tolerate incomplete descriptions and mid-process human correction.

\paragraph{The Inadaptability of Traditional Rule Engines}
Traditional software solutions are ill-equipped for this environment because they rely on rigid, pre-defined schemas. Since pricing conditions are unlimited, adding a new policy often requires more than just data entry; it necessitates a coordinated update of the \emph{database structure} (to store new attributes), the \emph{rule logic} (to handle new predicates), and the \emph{input UI} (to allow users to select these new options). This ``hard-coding'' cycle creates a massive maintenance burden, as software developers must constantly translate natural-language nuances into executable code. For large tourism sites with hundreds of evolving policies, the delay and cost of these manual system updates make traditional rule engines fundamentally unscalable.

\paragraph{Implications for System Design}
These constraints dictate that a production system cannot rely on a ``closed-world'' assumption. Instead, the system must: (i) decouple policy logic from the underlying software schema to handle an open-ended condition space, and (ii) externalize its reasoning into \emph{interpretable intermediate artifacts} that preserve natural-language semantics and treat human intervention as a first-class operation. 

\section{Methodology}
\label{sec:method}

We design an auditable pricing pipeline that operationalizes informal travel orders and natural-language pricing policies under strict governance constraints. The central design principle is a \emph{strict LLMs decision boundary}: LLMs are restricted to structured information extraction and discrete decision selection, while all numeric calculations are performed by a deterministic engine. This boundary structurally eliminates numeric hallucination, makes remaining uncertainty explicit, and enables rapid human verification.

\begin{figure*}[t]
  \centering
  \includegraphics[width=\textwidth]{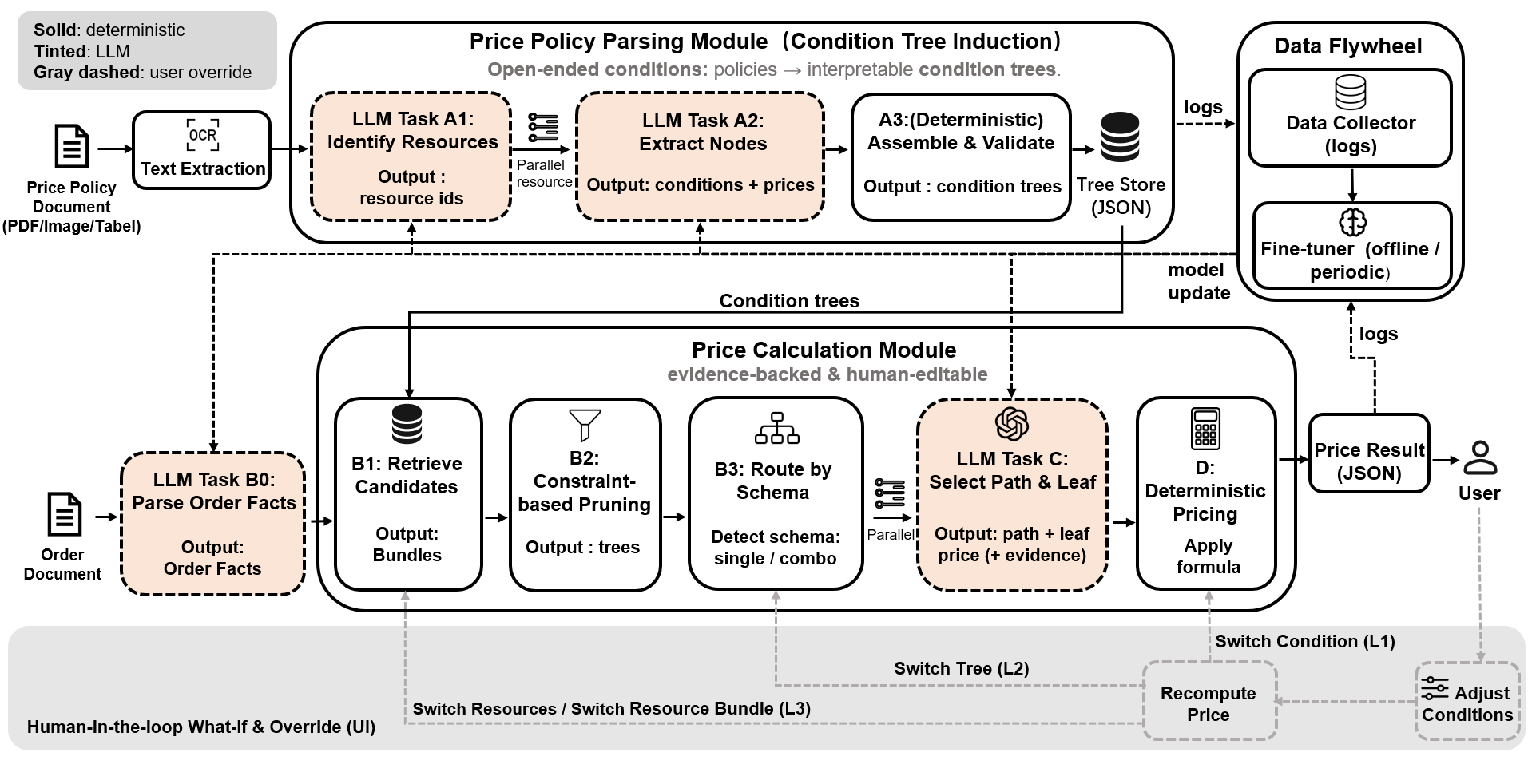}
  \caption{\small System overview. The policy parsing module induces interpretable condition trees from pricing documents, and the price calculation module routes orders to candidate trees, verifies applicability, and computes final prices with deterministic execution and human-in-the-loop overrides.}
  \label{fig:overview}
\end{figure*}

\begin{figure}[t]
  \centering
  \includegraphics[width=\linewidth]{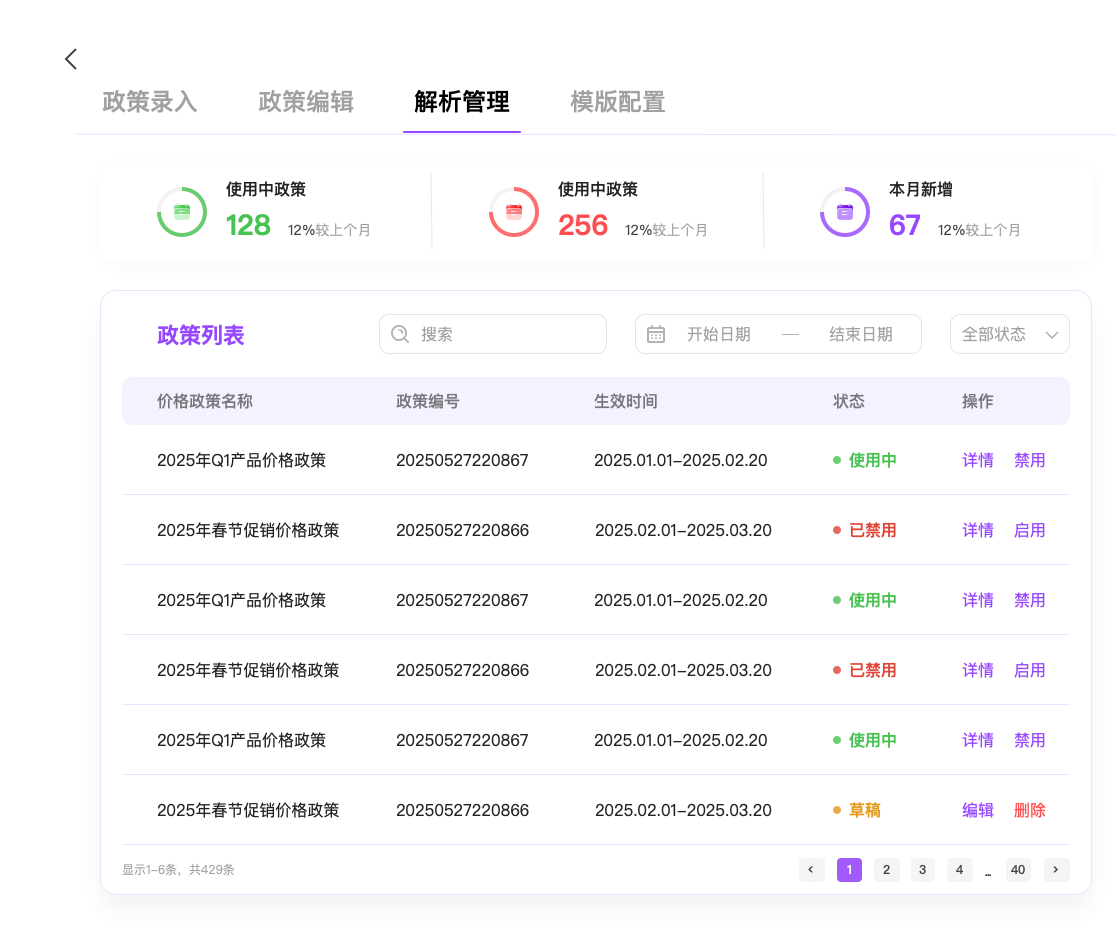}
  \caption{\small Policy management UI for validity windows and status, which serve as hard constraints for metadata-based routing (B2--B3).}
  \label{fig:ui_policy_manage}
\end{figure}

We provide the executable tree schema (Table~\ref{tab:full_schema}), the validator suite (Table~\ref{tab:validators}),
a production-faithful tree instance (Fig.~\ref{fig:tree_example}),
and end-to-end UI traces that externalize L1--L3 confirmations and overrides (Figs.~\ref{fig:ui_resource_edit}--\ref{fig:ui_price_result}). Before publication, every induced tree must pass an explicit validator suite (Table~\ref{tab:validators}), and any failure is blocked from entering execution and routed to manual correction; moreover, all tree management is presented to operators through a dedicated UI (Fig.~\ref{fig:ui_policy_manage}) for friendly oversight.
The system is decomposed into two major components: a \emph{policy onboarding module} (stages \emph{A0--A3}) and a \emph{price calculation module} (stages \emph{B0--D}). To ensure reliability, we incorporate a tiered human-in-the-loop override process (\emph{L1--L3}) that allows operators to intervene at distinct levels of the reasoning chain. These identifiers correspond to the architectural components illustrated in Figure~\ref{fig:overview}.

\subsection{Policy Onboarding and Tree Induction}
\label{sec:policy_parsing}

The policy processing module is the "offline" stage where policy documents are transformed into \emph{condition trees}. A condition tree acts as the sole executable interface for pricing a specific product. In this structure, each internal node represents a natural-language applicability condition, while each root-to-leaf path corresponds to exactly one \emph{price specification}. This \emph{one-path--one-price-spec} invariant guarantees that once a path is selected, the deterministic calculation.
The induction of these trees follows a four-step pipeline:

\begin{figure}[t]
  \centering
  \includegraphics[width=\linewidth]{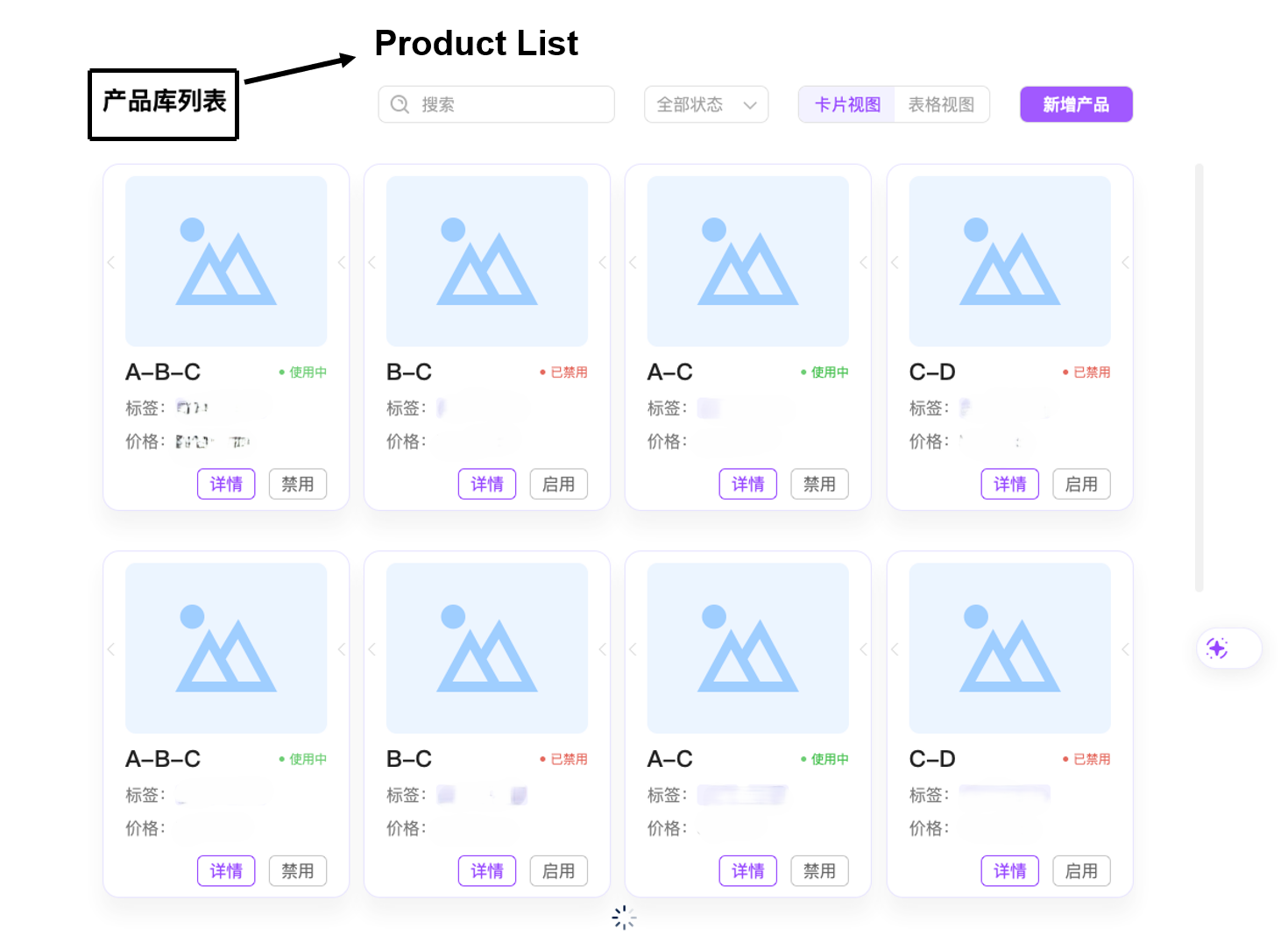}
  \caption{\small Catalog management UI defining the executable resource space $R$ (A1).}
  \label{fig:ui_catalog_R}
\end{figure}

\begin{figure}[t]
  \centering
  \includegraphics[width=\linewidth]{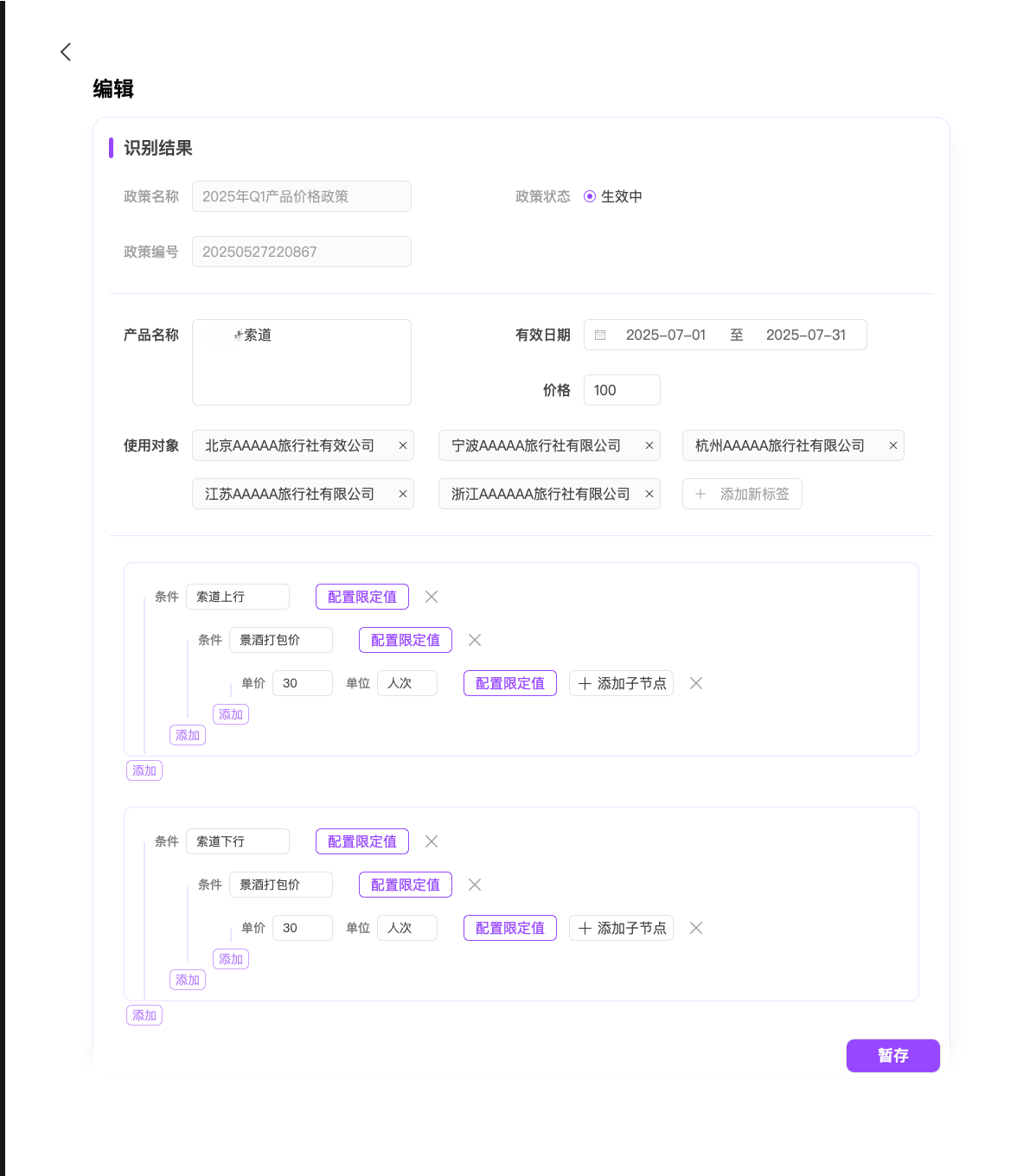}
  \caption{\small Policy parsing/editing UI for inducing and reviewing condition trees before publication (A2--A3).}
  \label{fig:ui_policy_parse}
\end{figure}

\noindent\textbf{A0: Text extraction.}
We first convert policy documents into text via OCR or table extraction, producing a normalized textual input.

\noindent\textbf{A1: Identify resources.}
We maintain a predefined resource catalog $R$ that serves as the canonical space of system-recognized resources. Each resource represents an operator-defined atomic capability that can be referenced by downstream execution components. In addition to defining individual resources, the catalog also specifies the product (bundle) configurations that can be constructed from these resources. As a result, $R$ constrains both the allowable \texttt{resource\_ids} and the feasible product combinations used in subsequent filtering steps (B1).
During policy interpretation, the LLM performs catalog grounding, mapping natural language policy descriptions into a set of resources drawn strictly from $R$. Because the grounding process operates over this predefined catalog, it prevents out-of-catalog entities from entering the executable interface. In our system architecture (Fig.~\ref{fig:ui_catalog_R}), the catalog $R$ is treated as a first-class operator artifact and serves as the only executable namespace for resources.
To ensure consistency of the catalog, uniqueness constraints are enforced during the catalog maintenance stage. At the resource level, each resource entry is uniquely identified by its name attribute in the database schema. Newly introduced resources are also subject to manual review to avoid semantic duplication or naming conflicts.
At the product level, each product represents a predefined bundle of resources stored in the catalog, and duplicate bundles are disallowed. During catalog maintenance, the system checks whether a candidate product definition duplicates an existing bundle. Given a list of resource names describing the candidate bundle, the system performs an exact matching query over the existing product–resource mappings in the database. The query verifies that the total number of resources matches and that the occurrence count of each resource is identical, while also ensuring that no additional resources are present. If such a product already exists, the candidate definition is rejected to prevent duplicate bundle entries.
These constraints ensure that the catalog remains consistent and unambiguous before it is used by the LLM grounding process. Consequently, all LLM operations rely exclusively on operator-defined and prevalidated catalog data, ensuring deterministic and reliable execution.

\begin{figure}[t]
  \centering
  \includegraphics[width=\linewidth]{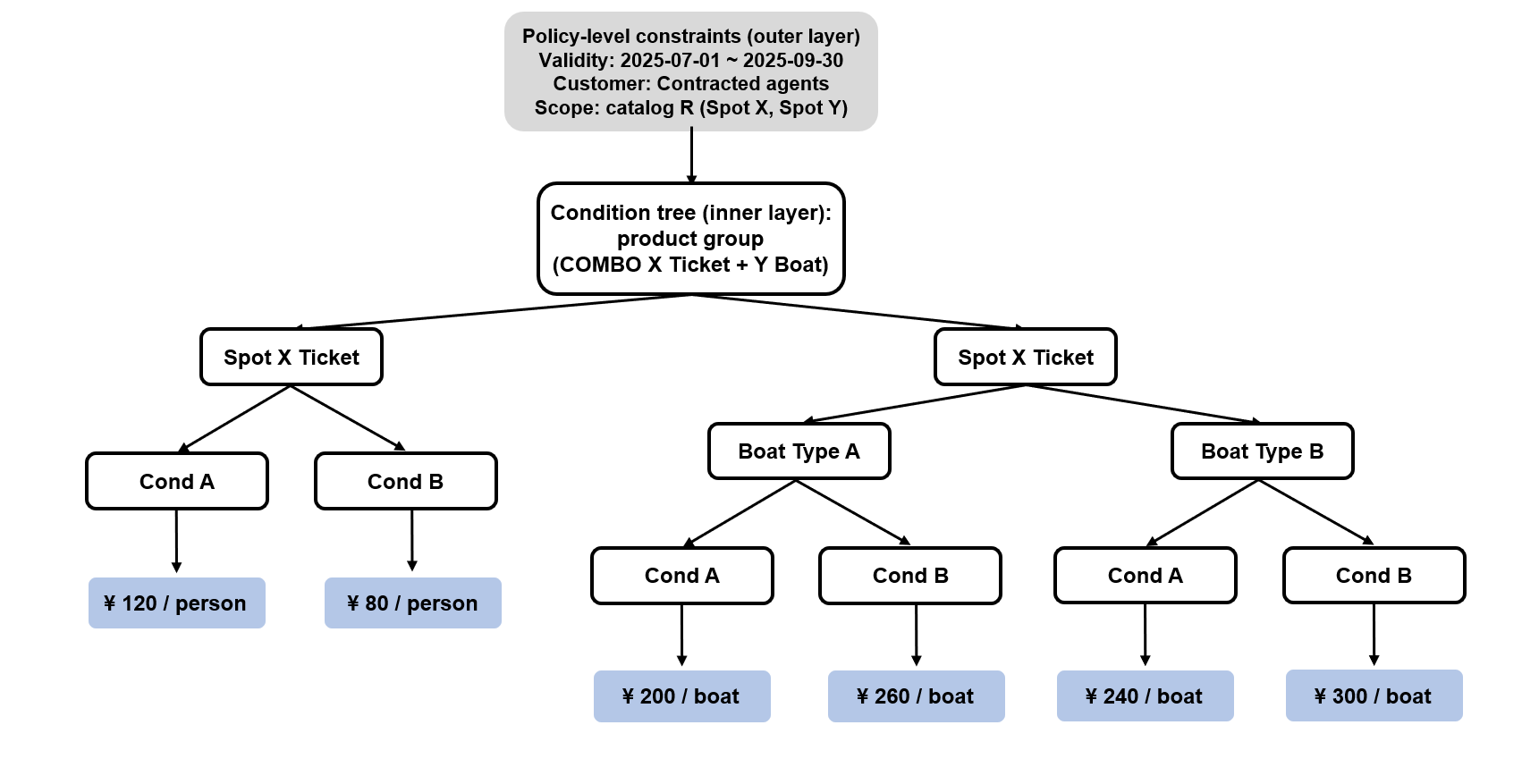}
  \caption{\small A production-faithful induced policy structure with an outer metadata layer and an arbitrarily nested condition tree.}
  \label{fig:tree_example}
\end{figure}

\noindent\textbf{A2: Extract nodes.}
For each grounded entity, the LLM extracts three types of information that align with the structure illustrated in Fig.~\ref{fig:tree_example}: (i) policy metadata, such as customer scope and validity windows (the outer metadata layer used for deterministic routing); (ii) the natural-language conditions that define the logic of the policy (the inner arbitrarily nested condition tree branching into resource-/product-specific subtrees); and (iii) the leaf-level \emph{price specifications}. Crucially, the LLM treats numeric values as static fields to be stored, rather than performing any calculations at this stage. Fig.~\ref{fig:tree_example} shows a production-faithful example of this structure, which motivates our choice of a UI-editable tree representation rather than hard-coded rule schemas.

\noindent\textbf{A3: Assemble \& validate.}
We assemble extracted clauses into an executable condition tree while enforcing the
\emph{one-path--one-price-spec} invariant: each root-to-leaf path corresponds to exactly one
leaf-level \emph{PriceSpec}, ensuring deterministic execution once a path is selected.
To make trees \emph{auditable} and \emph{UI-editable} in production, each node follows a
typed schema that separates (i) structural fields (\texttt{id}, \texttt{children}, \texttt{isLeafNode}),
(ii) semantic payload (\texttt{fieldName}, \texttt{fieldValue}), and (iii) leaf-only pricing fields
(\texttt{price}, \texttt{unit}), plus constraint/UI metadata such as \texttt{limitValue} and \texttt{expanded}
(Table~\ref{tab:full_schema}).
In particular, \texttt{limitValue} encodes applicability constraints as a two-level boolean form
(\textsc{and}/\textsc{or} over groups, each group over atomic predicates), which is expressive enough
for our policy space while remaining easy to validate and render consistently.

Before publication, every induced tree must pass an explicit validator suite
(Table~\ref{tab:validators}) that checks schema completeness, unit compatibility, numeric sanity,
catalog resolvability, constraint-format compliance, and cross-path conflict detection under the same
metadata window. Any failure blocks the tree from entering execution and routes it to operator review.
Operators inspect and edit the induced structure in an onboarding console
(Fig.~\ref{fig:ui_policy_parse}), where the nested condition hierarchy and leaf PriceSpecs are displayed
in a tree form, making extraction errors observable and correctable prior to release.
Only validated trees are published to the \emph{Tree Store}.

\begin{table}[t]
\centering
\small
\setlength{\tabcolsep}{4pt}
\renewcommand{\arraystretch}{1.12}
\begin{tabular}{@{}llp{0.50\linewidth}@{}}
\toprule
\textbf{Field} & \textbf{Type} & \textbf{Description} \\
\midrule
\texttt{id}          & int         & Unique node identifier \\
\texttt{fieldName}   & str         & Node category (e.g., product type, condition) \\
\texttt{fieldValue}  & str         & Category value (product name or condition clause) \\
\texttt{children}    & array       & Child nodes; empty array for leaves \\
\texttt{isLeafNode}  & bool        & Whether this node is a leaf \\
\midrule
\multicolumn{3}{@{}l}{\textit{Leaf-only fields (pricing specification)}} \\
\texttt{price}       & number      & Unit price \\
\texttt{unit}        & str/null    & Billing unit (e.g., per person, per group) \\
\midrule
\multicolumn{3}{@{}l}{\textit{Constraint and UI fields}} \\
\texttt{limitValue}  & object/null & Structured applicability constraints \\
\texttt{limit}       & str         & Reserved for future use \\
\texttt{expanded}    & bool        & UI expansion state (default: true) \\
\bottomrule
\end{tabular}
\caption{Full schema of condition-tree nodes. Fields are grouped by function: identification and structure (top), leaf pricing specification (middle), and constraints/UI metadata (bottom).}
\label{tab:full_schema}
\end{table}

\begin{table}[t]
\centering
\small
\setlength{\tabcolsep}{4pt}
\renewcommand{\arraystretch}{1.12}
\begin{tabular}{@{}p{0.32\linewidth}p{0.60\linewidth}@{}}
\toprule
\textbf{Validator} & \textbf{Purpose} \\
\midrule
\texttt{schema\_complete}     & All required fields present; tree is well-formed \\
\texttt{unit\_compat}         & Billing units are consistent and compatible across sibling leaves \\
\texttt{numeric\_sanity}      & Prices are non-negative and within plausible bounds \\
\texttt{conflict\_detect}     & No contradictory rules under overlapping metadata (scope, validity) \\
\texttt{service\_name}        & Resource/service names resolve to valid catalog entries \\
\texttt{restrict\_format}     & \texttt{limitValue} structure conforms to the two-level boolean schema \\
\bottomrule
\end{tabular}
\caption{Validators applied during policy onboarding. A tree is published only if all validators pass.}
\label{tab:validators}
\end{table}

\begin{figure}[t]
  \centering
  \includegraphics[width=\linewidth]{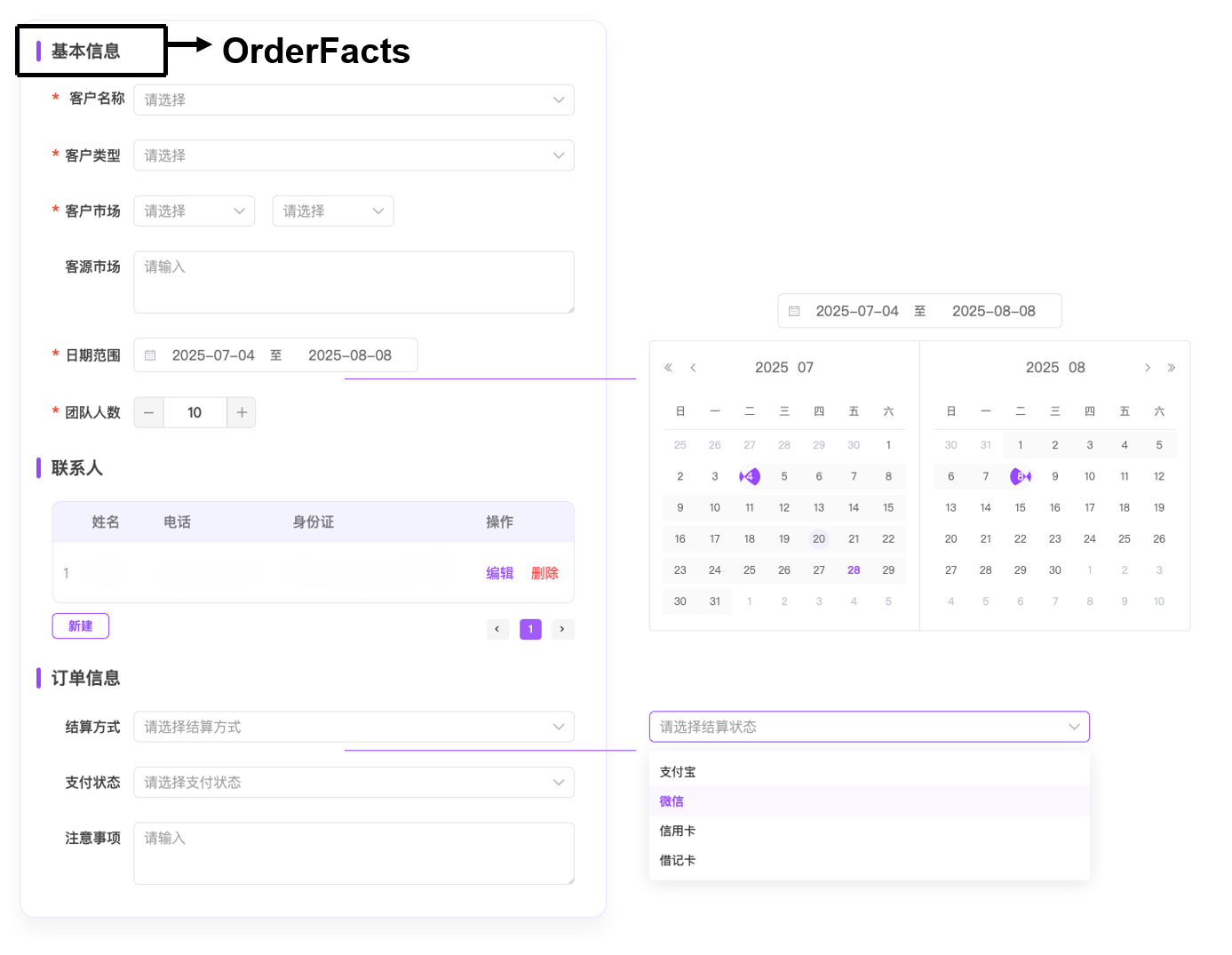}
  \caption{\small OrderFacts panel for order intake. Extracted structured fields are visible and editable before downstream bundle enumeration and policy matching (B0).}
  \label{fig:ui_orderfacts}
\end{figure}

\subsection{Order Parsing and Price Calculation}
\label{sec:price_calc}

\begin{figure}[t]
  \centering
  \includegraphics[width=\linewidth]{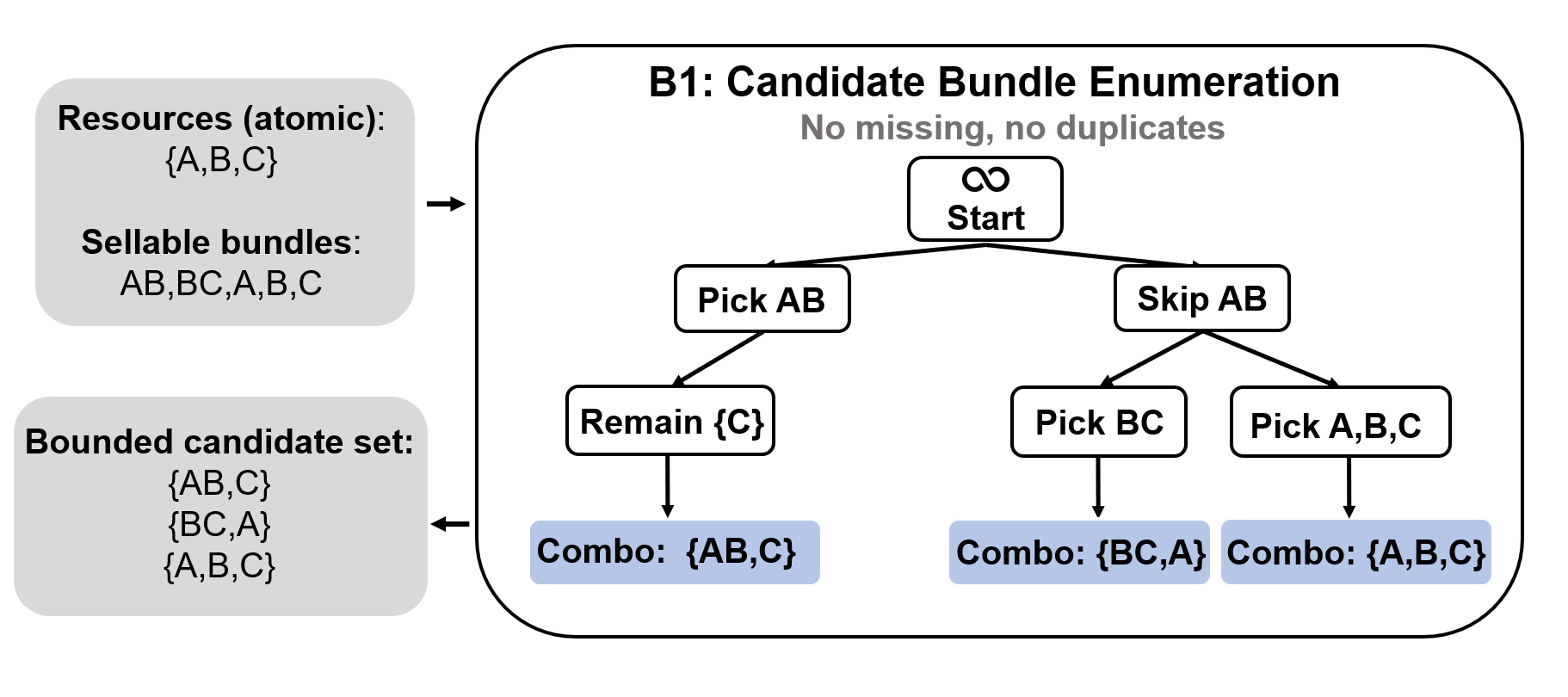}
  \caption{\small \textbf{B1: Candidate bundle enumeration.} Given atomic resource mentions and a catalog of sellable bundles, we deterministically enumerate all feasible bundle decompositions that exactly exhaust the requested resources for downstream tree routing and path selection.}
  \label{fig:bundle_enum}
\end{figure}

\begin{figure}[t]
  \centering
  \includegraphics[width=\linewidth]{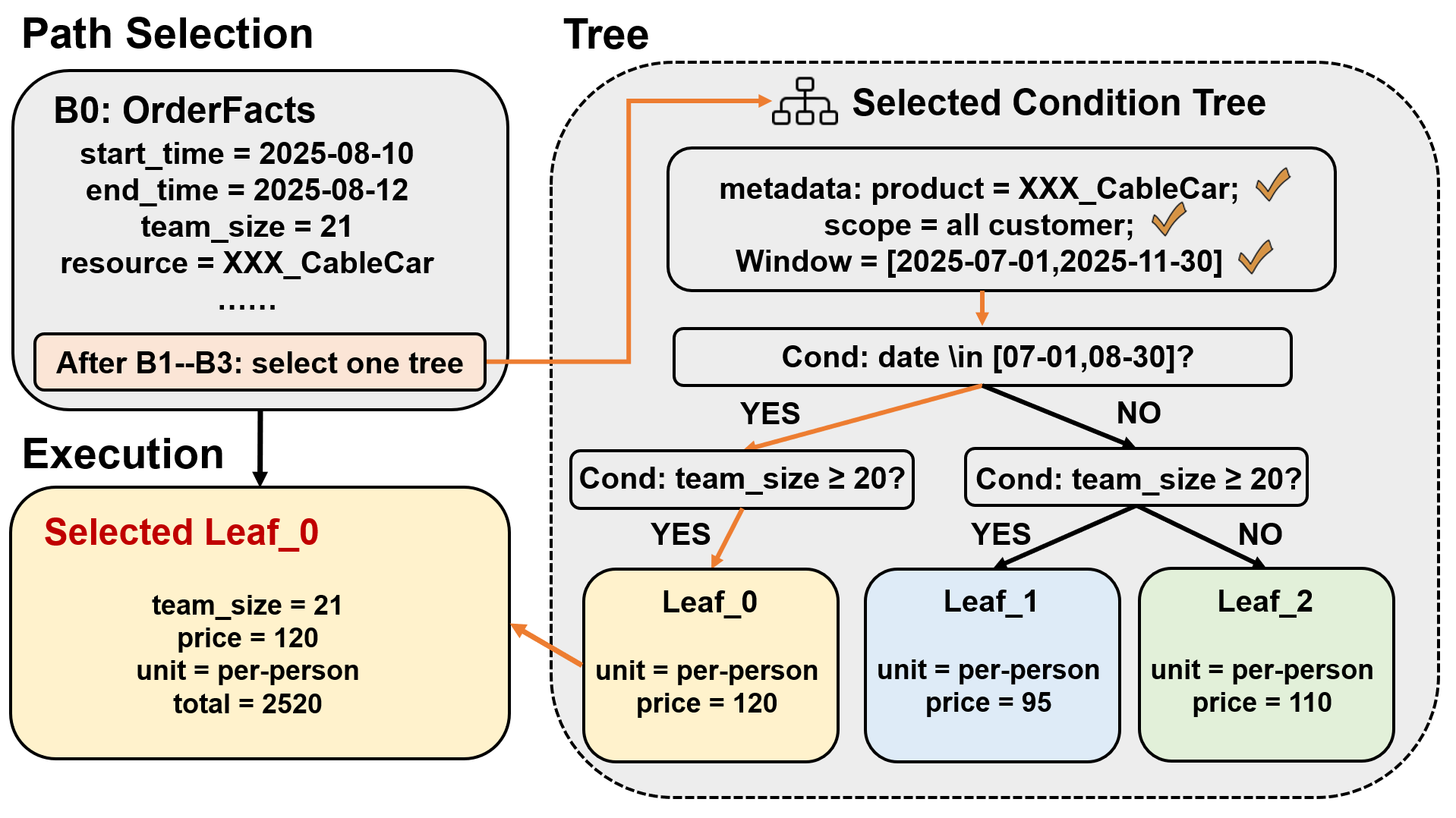}
  \captionsetup{font=small}
  \caption{\textbf{Toy walkthrough of tree-guided pricing.}
  The LLM extracts \emph{OrderFacts} (B0).
  B1--B3 deterministically route the order to a \emph{bounded} policy tree via metadata filtering.
  The LLM then performs \emph{discrete} root-to-leaf path selection (C) and returns a leaf-level \emph{PriceSpec},
  while the deterministic engine computes the final amount from validated quantities (D).
  A production-faithful condition tree with explicit metadata constraints is shown in Fig.~\ref{fig:tree_example}.}
  \label{fig:demo_v6}
\end{figure}

Once policies are onboarded as condition trees, the price calculation module processes incoming orders through the following stages:

\noindent\textbf{B0: Order Parsing and Fact Extraction}
Incoming orders (typically screenshots) are parsed by an LLM into \emph{OrderFacts} (\emph{B0}). This structured record contains non-resource fields, such as travel dates, group size, and customer identity, and grounded resource mentions.OrderFacts are surfaced in the operator console as an editable, structured form to correct missing/ambiguous fields before downstream routing and execution (Fig.~\ref{fig:ui_orderfacts}). This editability is essential because many screenshot orders are under-specified or later revised:
operators can correct missing dates/team size and reconcile resource mentions before deterministic
routing and execution, turning ambiguity into an explicit, auditable intervention rather than a silent failure.

\noindent\textbf{B1: Candidate Bundle Enumeration}
Since orders may mention individual resources that are sold as \emph{products} (bundles), the system must determine the best way to group these resources. We use a deterministic DFS backtracking algorithm to enumerate all feasible bundle combinations that exactly exhaust the requested resources; Figure~\ref{fig:bundle_enum} gives a toy illustration.

\noindent\textbf{B2--B3: Retrieval and Routing}
For each candidate bundle/product enumerated in B1, we retrieve its associated condition trees from the \emph{Tree Store} and deterministically filter/prune them using policy metadata constraints, including: (i) overlap between the order date range and the policy validity window, and (ii) match between the order-side customer scope and the policy applicability scope. After obtaining a bounded set of viable trees per candidate bundle/product, we further route candidates deterministically by a predefined pricing schema (e.g., \textit{single-item} vs.\ \textit{bundle} structures) to select the appropriate matching policy trees. In production, routing metadata (validity window, customer scope, and policy status) is governed as a
first-class artifact: policies are versioned, activated/deactivated, and audited through an operator console
(Fig.~\ref{fig:ui_policy_manage}). This ensures that deterministic routing only considers policies that are
currently effective and approved, preventing stale or unofficial rules from entering execution.

\noindent\textbf{C--D: Path Selection and Execution}
As shown in Figure~\ref{fig:demo_v6}, in the final reasoning step, an LLM analyzes the filtered trees and selects the root-to-leaf path whose conditions are satisfied by the \emph{OrderFacts} (\emph{C}); in production, the same selection operates over policy trees with an explicit metadata layer and richer nested subtrees (Fig.~\ref{fig:tree_example}). To ensure transparency, the model provides evidence pointers to specific spans in the order text. Once a path is confirmed, the deterministic engine executes the final pricing (\emph{D}) by applying the quantity normalization and units defined in the selected \emph{PriceSpec}.
The PriceResult view externalizes the selected candidate tree (L2) and matched path/leaf (L1) together with resource-level breakdown and recomputable totals, enabling auditable confirmation and bounded overrides within the exposed candidate set (Fig.~\ref{fig:ui_price_result}).

\subsection{Governance and Self-Evolving}
\label{sec:overrides}
\paragraph{Override Protocol}

\begin{figure}[t]
  \centering
  \includegraphics[width=\linewidth]{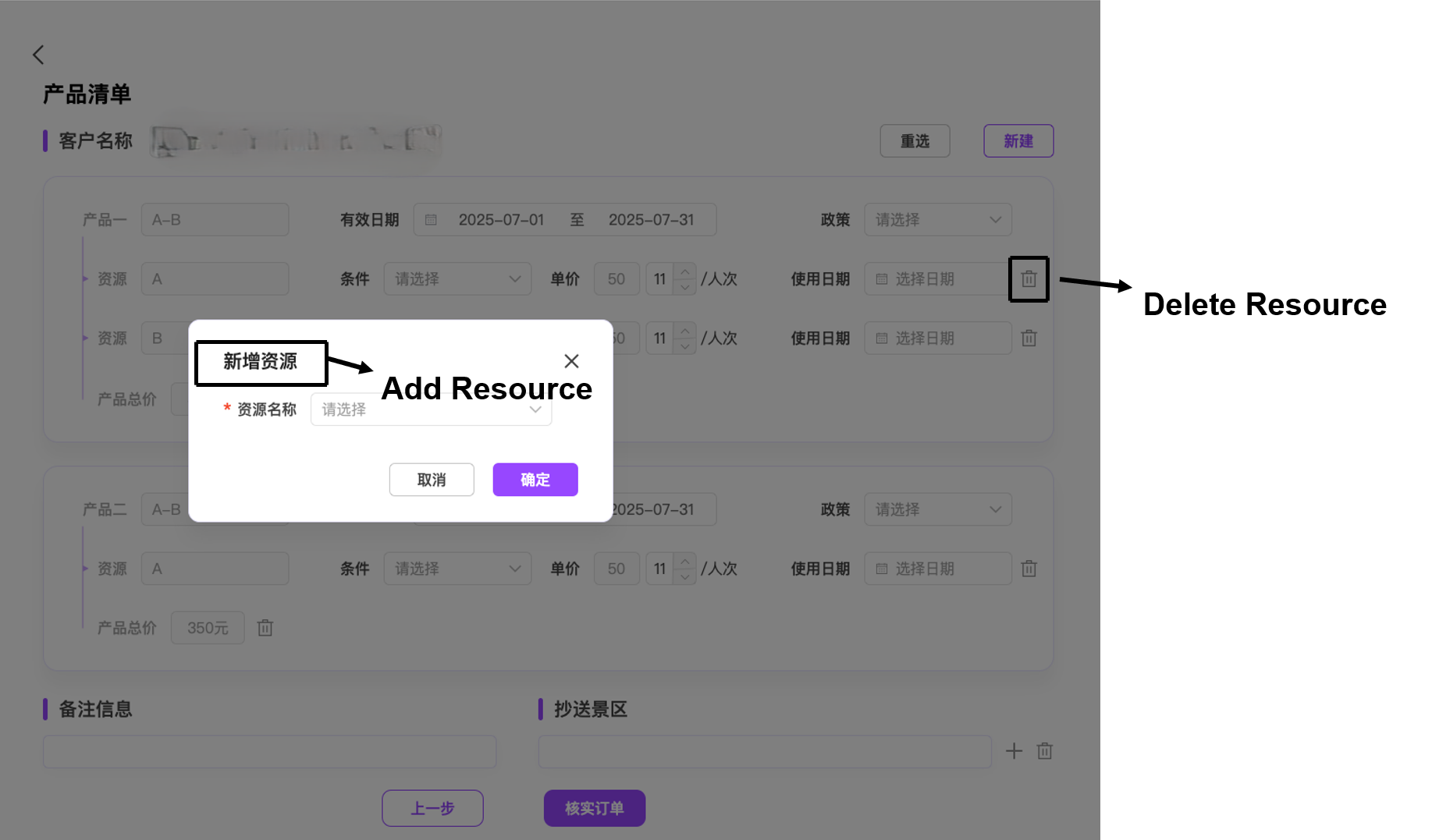}
  \caption{\small Add/remove resource UI (\emph{L3-a resource edit}). Operators reconcile extracted resources with the actually executed itinerary before pricing.}
  \label{fig:ui_resource_edit}
\end{figure}

\begin{figure}[t]
  \centering
  \includegraphics[width=\linewidth]{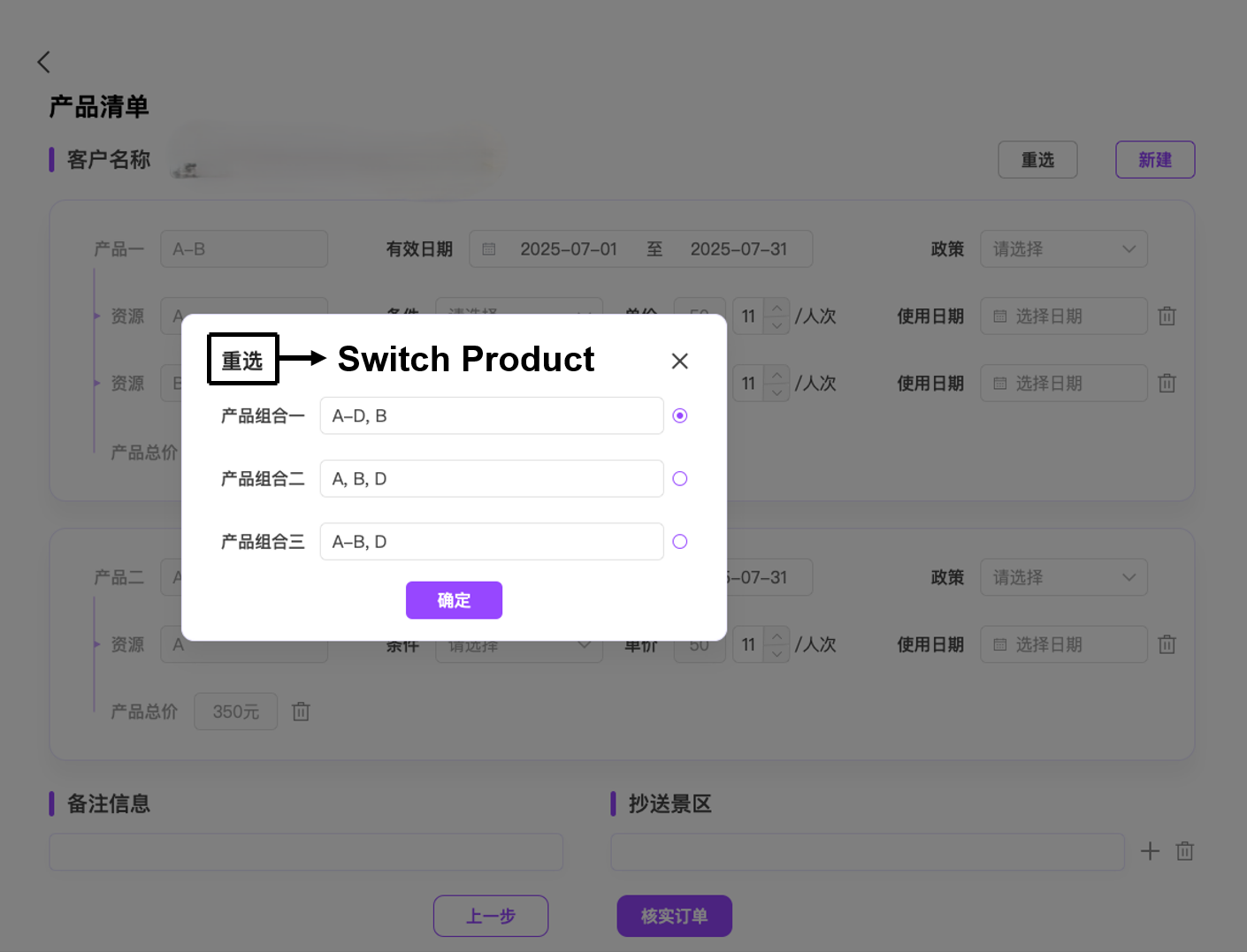}
  \caption{\small Bundle/product switching UI (\emph{L3-b composition override}). Operators switch among feasible bundle decompositions enumerated in B1.}
  \label{fig:ui_switch_bundle}
\end{figure}

\begin{figure}[t]
  \centering
  \includegraphics[width=\linewidth]{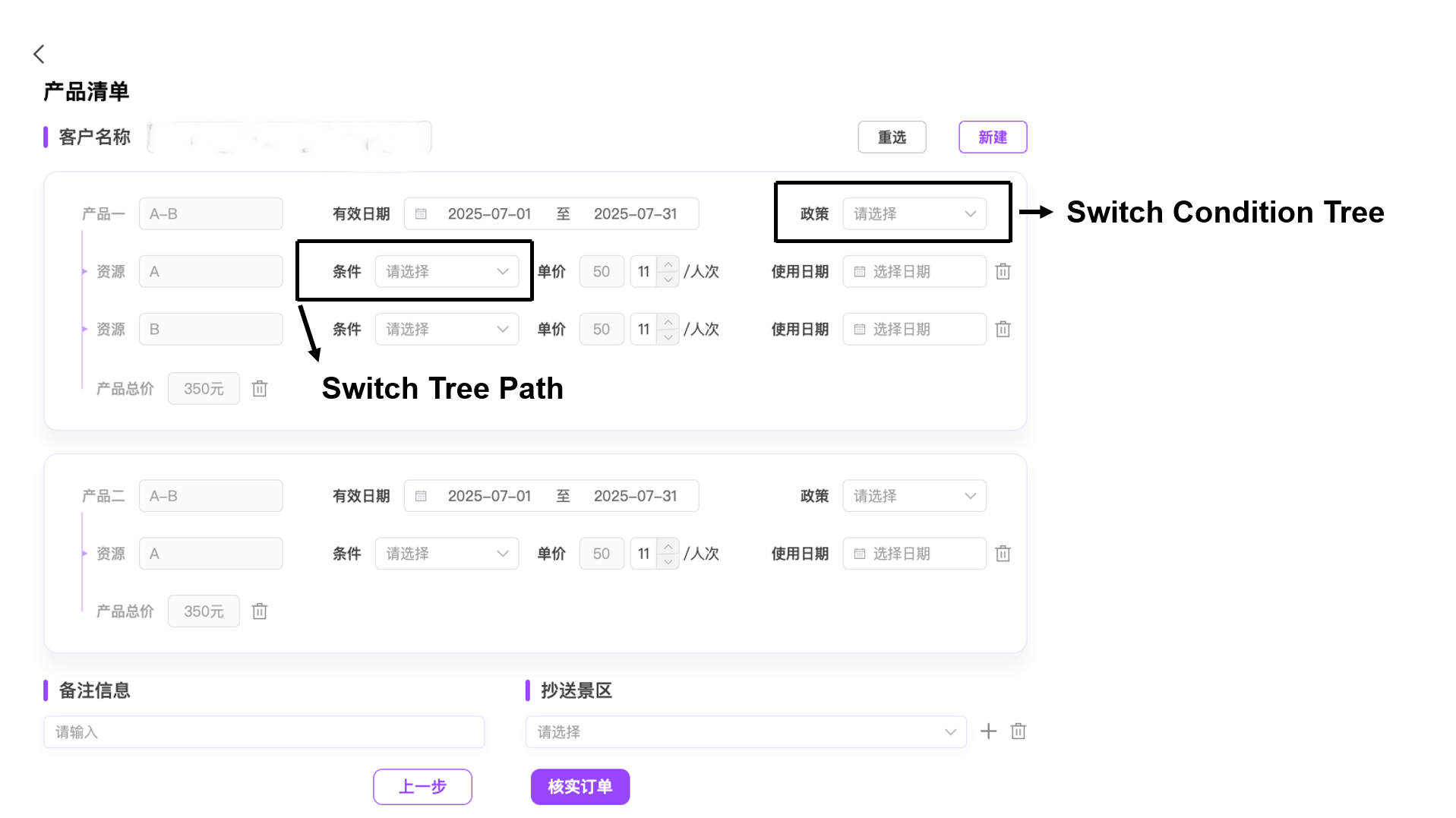}
  \caption{\small PriceResult view. Each item externalizes the selected tree (\emph{L2}) and matched path/leaf (\emph{L1}) with recomputable breakdown; operators can switch the candidate tree (L2) or candidate path (L1) within the exposed set.}
  \label{fig:ui_price_result}
\end{figure}

To maintain zero-tolerance for numeric errors, we expose the system's reasoning through three override levels:
(1) \emph{L1 (Path Override):} The operator selects a different branch within a fixed tree, correcting LLM errors of natural-language conditions (Fig.~\ref{fig:ui_price_result}).
(2) \emph{L2 (Tree Override):} The operator switches between valid policy trees, typically used to resolve overlapping policy conflicts (Fig.~\ref{fig:ui_price_result}).
(3) \emph{L3 (Composition/Data Override):} The operator corrects upstream inputs, either by editing the
resource list (\emph{L3-a}) to fix extraction or itinerary mismatch (Fig.~\ref{fig:ui_resource_edit}),
or by switching the bundle decomposition/product (\emph{L3-b}) when composition differs from the
actual itinerary (Fig.~\ref{fig:ui_switch_bundle}).

\paragraph{Failure Modes and Data Flywheel}

\begin{figure}[t]
  \centering
  \includegraphics[width=\linewidth]{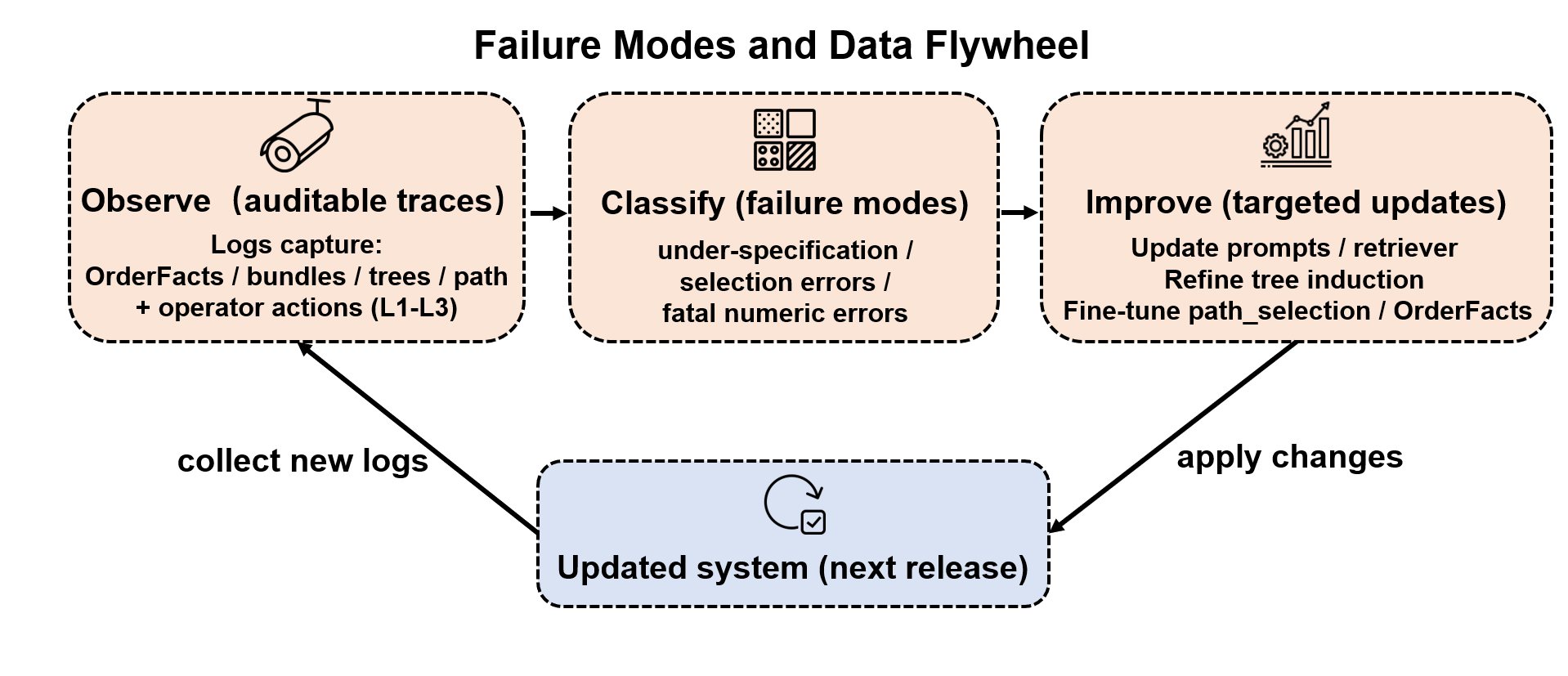}
  \caption{\small \textbf{\small Failure-mode taxonomy and data flywheel.}}
  \label{fig:flywheel}
\end{figure}

Production failures are categorized into three types: \emph{under-specification} (incomplete order info), \emph{selection errors} (incorrect path choice), and \emph{fatal numeric errors}. As shown in Figure~\ref{fig:flywheel}, our architecture is designed to make the first two types recoverable via the \emph{L1--L3} protocols, while the third is prevented by our strict decision boundary.
Finally, we maintain a \emph{production data flywheel} by logging raw model outputs alongside human-confirmed overrides. This allows for iterative improvement of extraction and selection prompts without risking the stability of the deterministic execution logic.

\section{Experiment}
\label{sec:exp}

We deploy our system on a city-level state-owned tourism enterprise. This enterprise is responsible for managing seven major scenic areas across the city, encompassing 12 diverse tourism business categories across land, water, and air dimensions (including accommodation, dining, car rental, attraction visits, shopping, entertainment, tour guide services, performances, cruise ships, conferencing, cable cars, drones, etc.), with over 1,500 operational staff, complex organizational structure, and diverse role distributions. We evaluate our system on production logs from 2025H2 (Jul--Dec 2025).
The dataset comprises 3,960 orders, of which 3,842 (97.0\%) are instant-messaging screenshots processed via OCR and 118 (3.0\%) are Word uploads.

We focus on screenshot orders as the dominant deployment modality.
Table~\ref{tab:data_complexity} summarizes order structural complexity: over 70\% of orders involve multiple resources, and 25\% contain bundle products, motivating the deterministic composition enumeration in stage B1.

\begin{table}[t]
\centering
\small
\setlength{\tabcolsep}{4pt}
\renewcommand{\arraystretch}{1.10}
\begin{tabular}{@{}lrr@{}}
\toprule
\textbf{Category} & \textbf{Count} & \textbf{\%} \\
\midrule
Single-resource orders     & 1{,}178 & 29.7 \\
Multi-resource orders      & 2{,}782 & 70.3 \\
\quad 2 resources          & 1{,}858 & 46.9 \\
\quad 3+ resources         &   924   & 23.4 \\
Orders with bundle products &  990   & 25.0 \\
\bottomrule
\end{tabular}
\caption{\small Order structural complexity (2025H2, $n=3{,}960$).}
\label{tab:data_complexity}
\end{table}

\paragraph{Evaluation Protocol}

We organize evaluation as a decision-boundary funnel that progressively isolates each decision level, so that downstream metrics reflect only the residual decision.
In Track~1, we quantify operator workload on all 3,842 screenshot orders. We distinguish two mutually exclusive L3 interventions: L3-a (resource edit) addresses mismatch between the extracted resource set and the actually executed itinerary, while L3-b (bundle/product switch) addresses composition mismatch when the resource set is already consistent. After removing all L3 cases, we measure L2 (tree switching) on the remaining orders.
In Track~2, we isolate within-tree path selection (L1) and pricing correctness. From the 1,785 orders that survive L2 and L3 filtering, we further restrict to an information-consistent subset where critical pricing fields (travel dates and team size) are consistent with logged outcomes, yielding 1,349 orders covering 2,598 matched product instances. All correctness metrics are reported on this subset.
We report three metrics aligned with production requirements. First, workload is measured by the frequency of L3-a, L3-b, and L2 interventions across the full screenshot set. Second, unit-price accuracy is the fraction of product instances whose predicted unit price matches any logged ground-truth price under the same order-product key. Third, the fatal numeric error rate captures any output with invalid units, aggregation mismatch, or prices unsupported by a verified leaf specification.

\paragraph{Results}

Table~\ref{tab:results_summary} presents the complete decision-boundary funnel.
\begin{table}[t]
\centering
\small
\setlength{\tabcolsep}{3pt}
\renewcommand{\arraystretch}{1.10}
\begin{tabular}{@{}p{0.52\linewidth}rr@{}}
\toprule
\textbf{Decision boundary} & \textbf{Count} & \textbf{Rate} \\
\midrule
\multicolumn{3}{@{}l}{\textit{Track 1: Production workload ($n{=}3{,}842$)}} \\
\quad L3-a resource edits        & 1{,}842 & 47.9\% \\
\quad L3-b bundle/product switch &   108   &  2.8\% \\
\quad No L3 intervention         & 1{,}892 &   —   \\
\quad L2 tree switching          &   107   &  5.7\% \\
\quad No L2/L3 intervention      & 1{,}785 &   —   \\
\midrule
\multicolumn{3}{@{}l}{\textit{Track 2: Correctness ($n{=}2{,}598$ product inst.)}} \\
\quad Unit correctness           & 2{,}598 / 2{,}598 & 100\% \\
\quad Unit-price accuracy (L1)   & 2{,}214 / 2{,}598 & 85.2\% \\
\quad Fatal numeric error   &     0 / 2{,}598   &   0\% \\
\bottomrule
\end{tabular}
\caption{Decision-boundary funnel. Track 1 reports operator workload on the full screenshot set; Track 2 reports correctness on the information-consistent subset where only L1 remains.}
\label{tab:results_summary}
\end{table}
On the full 3,842 screenshot orders, 1,842 (47.9\%) require L3-a resource edits and 108 (2.8\%) require L3-b bundle/product switching. The high L3-a rate reflects a known operational reality: coordinators frequently omit or revise resources after submission, so this is not a system error but a characteristic of the domain workflow. Among the 1,892 orders requiring no composition-level intervention, only 107 (5.7\%) require L2 tree switching, indicating that metadata-based routing resolves the vast majority of policy-selection decisions automatically.
On the 1,349 information-consistent orders (2,598 product instances), the system achieves zero fatal numeric errors, confirming that the strict LLM decision boundary---where the model selects discrete paths rather than generating prices---structurally eliminates numeric hallucination. Unit-price accuracy reaches 85.2\%; the remaining 14.8\% are L1 path-selection errors that operators can recover via a single UI click rather than requiring any change to the execution layer.

\paragraph{Operational Impact}

Table~\ref{tab:operational} summarizes key deployment metrics collected over one year of production operation at a tourism site comprising 19+ scenic spots and over 1,000 annual pricing policies.
\begin{table}[t]
\centering
\small
\setlength{\tabcolsep}{4pt}
\renewcommand{\arraystretch}{1.10}
\begin{tabular}{@{}lcc@{}}
\toprule
\textbf{Metric} & \textbf{Before} & \textbf{After} \\
\midrule
Policy onboarding time & ${\sim}$20 min & ${\sim}$1 min \\
Order processing time  & manual & ${\sim}$2 min \\
Fatal numeric errors   &   —   &  0\% \\
Developer involvement  & required & not required \\
\bottomrule
\end{tabular}
\caption{\small Operational impact before and after deployment.}
\label{tab:operational}
\end{table}
The most significant operational gain is the elimination of developer involvement in policy onboarding. Previously, each new pricing policy required an expert developer to translate natural-language conditions into executable rules, taking approximately twenty minutes per policy. With our system, non-technical tourism managers can review and publish condition trees directly, reducing onboarding time to roughly one minute and eliminating the communication overhead between business and engineering teams. Order processing time is also reduced to about three minutes on average, covering the full pipeline from OCR extraction through pricing execution and operator confirmation.

\section{Discussion}
\label{sec:discussion}

The central design choice of our system is confining the LLM to discrete selection tasks---resource grounding, condition extraction, and path selection---while delegating all numeric computation to deterministic execution from validated leaf specifications. This strict decision boundary yields two concrete benefits observed in production. First, the zero fatal numeric error rate across 2,598 product instances demonstrates that preventing the LLM from generating prices structurally eliminates numeric hallucination. Second, all remaining errors are discrete and recoverable: operators correct path-selection mistakes via L1--L3 overrides without touching the execution layer, making the system's failure modes transparent and manageable.

The condition trees serve not merely as post-hoc explanations but as the primary editing interface for non-technical operators. This design is validated by the low L2 override rate (5.7\%): once a tree is correctly constructed during onboarding, metadata-based routing reliably selects it, and operators need only inspect and occasionally adjust the within-tree condition path. The interpretability-as-interface principle also simplifies the data flywheel---disagreements between model outputs and operator-confirmed results can be directly attributed to specific decision boundaries, enabling targeted iteration rather than opaque end-to-end retraining.

The 14.8\% unit-price mismatch in Track~2 is concentrated in two categories. 
The first is under-specification induced by open-ended eligibility conditions: a group may simultaneously satisfy multiple policy paths, and some discounted clauses require extra evidence that is typically unavailable in the order text (e.g., military ID, local employment/eligibility certificates). 
As a result, the correct unit price is not uniquely identifiable from \texttt{OrderFacts} alone; the system may select a verifiable default path while the logged outcome reflects additional offline information, leading to an apparent mismatch. 
The second is OCR noise from screenshot orders, where garbled text leads to incorrect \texttt{OrderFacts} extraction. 
Both are addressable without architectural changes: the former by making eligibility evidence explicit during onboarding and surfacing such paths as ``needs additional proof'' for operator confirmation, and the latter through improved OCR preprocessing or encouraging structured order input formats.

While our evaluation is conducted in tourism pricing, the framework's core components---condition tree induction, deterministic candidate enumeration, and evidence-backed discrete selection---are domain-agnostic. Industries with evolving natural-language policies and zero-tolerance pricing requirements, such as insurance underwriting, regulatory compliance, and contract execution, can adopt the same architecture by replacing the resource catalog and product dictionary with domain-specific counterparts.

\section{Conclusion}
\label{sec:conclusion}

We present a production LLM framework that converts natural-language pricing policies into executable, interpretable condition trees and processes informal travel orders through a deterministic pricing pipeline. By strictly separating LLM-based discrete selection from deterministic numeric execution, the system achieves zero fatal pricing errors across thousands of production orders, while reducing policy onboarding time from twenty minutes to one minute without developer involvement. In a six-month deployment at a municipal, state-owned tourism enterprise, the system processed 3{,}960 orders and delivered substantial operational gains, demonstrating that LLMs can be safely integrated into high-stakes decision workflows when paired with structured, auditable reasoning interfaces. The framework generalizes to other policy-driven domains where interpretability, reliability, and non-expert usability are equally critical.

\section*{Limitations}
\label{sec:limitations}
Our current deployment relies on a manual grouping strategy that merges policies with similar customer scope and validity patterns into shared condition trees, in order to limit the number of trees per product. As the system runs longer and policies accumulate, this grouping may degrade, increasing matching ambiguity and operator burden. Developing automatic tree merging and splitting strategies is a natural direction for future work.

Nearly half of production orders (47.9\%) require L3-a resource edits, reflecting the domain-specific gap between the resources stated in the order and the itinerary actually executed. While the system surfaces this mismatch for operator correction, reducing this rate---through richer order input formats or multi-turn clarification with coordinators---would further lower the human workload.

Our correctness metrics in Track~2 are computed on an information-consistent subset of 1,349 orders where composition and tree selection are pre-resolved. End-to-end accuracy on the full order set, including undecidable cases where the input is genuinely ambiguous, remains difficult to evaluate without additional ground-truth annotations. Furthermore, all experiments are conducted at a single tourism deployment site. Although the architecture is designed to be domain-agnostic, empirical validation in other policy-driven domains such as insurance or regulatory compliance is needed to confirm generalizability.

\bibliography{custom}

\end{document}